\documentclass[conference]{IEEEtran}
\IEEEoverridecommandlockouts
\usepackage{cite}
\usepackage{amsmath,amssymb,amsfonts}
\usepackage{algorithmic}
\usepackage{textcomp}
\usepackage{xcolor}
\usepackage{fancyhdr}
\usepackage{indentfirst} 
\usepackage{float}

\usepackage{lipsum}
\makeatletter
\def\footnoterule{\relax%
  \kern 5pt
  \hbox to \columnwidth{\hfill\vrule width 0.95\columnwidth height 0.4pt\hfill}
  \kern 4pt}
\makeatother

\usepackage{graphicx}
\usepackage{subcaption}
\usepackage[utf8]{inputenc}
\usepackage[export]{adjustbox}
\usepackage{wrapfig}

\usepackage{multirow}

\usepackage{booktabs}
\usepackage{array}
\usepackage{enumerate}
\usepackage{tabularx,booktabs}

\AtBeginDocument{%
  \providecommand\BibTeX{{%
    \normalfont B\kern-0.5em{\scshape i\kern-0.25em b}\kern-0.8em\TeX}}}

\begin{document}

\title{Insight from NLP Analysis:  COVID-19 Vaccines Sentiments on Social Media}
\author{
	\IEEEauthorblockN{Tao Na\IEEEauthorrefmark{1}, Wei Cheng\IEEEauthorrefmark{1}, Dongming Li\IEEEauthorrefmark{1}, Wanyu Lu\IEEEauthorrefmark{1}, Hongjiang Li\IEEEauthorrefmark{1}}
	\IEEEauthorblockA{\IEEEauthorrefmark{1}
		Department of Computer Science, University of Manchester, Manchester, UK
		\\Correspondence to: Tao Na 
		\textless tao.na@student.manchester.ac.uk\textgreater \\
		Each author has equal contribution}
}

\maketitle
\thispagestyle{fancy}
\cfoot{\thepage} 
\pagestyle{fancy}
\cfoot{\thepage} 
\renewcommand{\headrulewidth}{0pt} 
\renewcommand{\footrulewidth}{0pt} 

\begin{abstract}
Social media is an appropriate source for analyzing public attitudes towards the COVID-19 vaccine and various brands. 
Nevertheless, there are few relevant studies. 

In the research, we collected tweet posts by the UK and US residents from the Twitter API during the pandemic and designed experiments to answer three main questions concerning vaccination.
To get the dominant sentiment of the civics, we performed sentiment analysis by VADER and proposed a new method that can count the individual's influence in. This allows us to go a step further in sentiment analysis and explain some of the fluctuations in the data changing. 
The results indicated that celebrities could lead the opinion shift on social media in vaccination progress. Moreover, at the peak, nearly 40\% of the population in both countries have a negative attitude towards COVID-19 vaccines. 
Besides, we investigated how people's opinions toward different vaccine brands are. We found that Pfizer vaccine enjoys the most popular among people. By applying the sentiment analysis tool, we discovered most people hold positive views toward the COVID-19 vaccine manufactured by most brands.
In the end, we carried out topic modelling by using the LDA model.
We found residents in the two countries are willing to share their views and feelings concerning the vaccine. 
Several death cases have occurred after vaccination. Due to these negative events, US residents are more worried about the side effects and safety of the vaccine.
\end{abstract}






\section{Introduction}
Since the first patient was identified in Wuhan, China, in December 2019\cite{zhu2020novel}, the COVID-19 has spread rapidly to Europe and eventually worldwide. COVID-19 can cause severe respiratory illness \cite{DBLP:journals/entropy/BajicDM21}. This complication has caused more than 2.17 million deaths. An outbreak of the virus has also been spreading in the UK and the US since March 2020. As of 22 March 2021, the number of confirmed cases in both countries exceeded 29.8 million and 4.3 million. COVID-19 has killed more than 668,000 people in the two countries. 

To prevent further spread of COVID-19 and relieve the enormous medical pressure, the development and promotion of vaccines are crucial. 
Several pharmaceutical companies and universities have been working on COVID-19 vaccines at an unprecedented rate. More than 260 possible COVID-19 vaccines have been proposed, but only a few have been approved. 
Several others are in state-of-the-art steps of testing\cite{VaccineCandidate}. Pfizer is the first international pharmaceutical company to have its vaccine approved in multiple countries: in the UK on 2 December 2020\cite{ledford2020uk}, in the US on 12 December 2020\cite{tanne2020covid} and in the EU on 21 December 2020\cite{Pfizer-Biontech}. 
At the time of writing, the availability of the COVID-19 vaccine is still limited, and less than 6\% of the global population has received the vaccine. As of 20 March 2021, vaccination coverage in the UK and US was 43.99\% and 36.31\%, respectively\cite{WorldData}.

The development of a safe vaccine through animal models of RSV can take up to thirty years\cite{graham2020rapid}. Vaccine development needs to be evaluated repeatedly in animal models before putting into clinical trials. Because of the severity of the epidemic, the development pace of new crown vaccines is unprecedented. Each country and global organization have lowered relevant criteria for new COVID-19 vaccines, and they have also shortened the clinical trials of vaccines\cite{JNJ}. Although the vaccines now in use have undergone rigorous testing and review to ensure safety, many people remain skeptical about the safety of the vaccines. There are small parts of people who even refused to be vaccinated with the COVID-19 vaccine. In order to promote the vaccine effectively, it is undoubtedly significant to collect people's opinions toward the vaccine and its side effects.
However, measures such as social isolation, quarantine and travel restrictions imposed by governments have hindered the collection process.
Therefore social media, like Twitter, becomes the ideal source of data.

To investigate the perceptions and attitudes of the UK and US citizens regarding the vaccine, we used the twitter API to collect relevant tweets during the outbreak. And we conducted social media analysis on these tweets. The dataset was collected by using the Twitter API. Based on a public dataset\footnote{Public dataset: https://zenodo.org/record/4603998\#.YGGK3GQzb0q} provided by Banda et al.\cite{banda2020large}, we only kept tweet\_id column of the English twitter posts with the location limitation to the US and the UK. And then we downloaded all COVID-19 vaccine-related tweets from the twitter stream via twitter API and the tweet ids. The public messages, like tweet\_id, date, time, lang and country\_place are included in our final dataset. The usage of dataset fully compliant with Twitter's terms of service.

Three main questions we analyze in this paper about COVID-19 vaccine and our contribution to each question are as shown below:
\begin{itemize}
  \item \textbf{What is the dominant sentiment towards COVID-19 vaccines? } For this question, We provide the analysis about the attitude of citizens locating in the UK and US toward the COVID-19 vaccine. The analysis is mainly conducted using a sentiment analysis tool, VADER\cite{hutto2014vader}, which is used primarily to explore the main sentiment expressed in people's tweets. In the research, we also proposed a new method that can capture the user’s public influence on the social network, thus contributing to the analysis of experiments result.
  
  \item \textbf{Which COVID-19 vaccine brands/manufacturers have been most talked about recently? Do people prefer any brands?}  Regarding to this, we explore the sentiment of people locating in the UK and US toward different COVID-19 vaccine brands. We manually determined vaccine brands and corresponding keywords that are currently talked most about on the Twitter platform. VADER is used to analyze people's preferences regarding different brands.
  
  \item \textbf{What people concern about the COVID-19 vaccine? What are the popular topics regarding vaccines?} With respect to this issue, we identified the main concerns among citizens about the COVID-19 vaccine. Using the LDA model, we explored the popular topics from the perspective of time series, country and sentiment, respectively.
\end{itemize}

%
\section{Related work}

Sentiment analysis, also called opinion mining, aims to evaluate embedded attitudes, opinions, sentiments, evaluations via the computational subjectivity in natural language text\cite{hutto2014vader}. Through sentiment analysis, we can know whether a text has a positive or negative subjective orientation.

Common sentiment analysis models rely heavily on sentiment dictionaries. A sentiment lexicon is a set of vocabulary in which each word is labelled according to the positivity and negativity of its subjective orientation.

Separately, the different lexicons can be classified into two types: semantic orientation labelling (divided into positive or negative) or more fine-grained quantitative scoring with predefined rules.
LIWC\cite{pennebaker2015development}, GI\cite{stone1966general}, HU-LIU04\cite{hu2004mining} are widely used polarity-based lexicons in which words are context-free. In contrast, ANEW\cite{bradley1999affective}, SentiWordNet\cite{baccianella2010sentiwordnet} and SenticNet\cite{cambria2012senticnet} are based on sentiment intensity thus could conduct a quantitative scoring evaluation.

The VADER (Valence Aware Dictionary for Sentiment Reasonable) proposed by C.J. Hutto et al.\cite{hutto2014vader} Is both a polarity and intensity Aware Dictionary. Especially, its performance in the field of social media text is exceptionally excellent. Based on its complete rules, VADER can carry out sentiment analysis on various lexical features: Punctuation, capitalization, degree modifiers, the contrastive conjunction "but", negation flipping tri-gram. Therefore, VADER could address the challenges stem from informal language usage and implicitly sentiment expression. 

Automated identifying sentiment features in the text, through the deep learning approach, is also a study direction. However, most of these methods are unstable and have some questions: First, such an approach requires a considerable amount of tagged emotional vocabulary data that is often challenging to obtain towards a particular text-domain. Second, the deep model is Computational extensive in training, validation and testing, and the model's performance in predicting tasks directly limit its ability to process streaming data. Third, such deep models are usually of the black box type, with limited interpretability.

Because of the advantages of VADER, it has a wide range of applications. Toni Pano and Rasha Kashef\cite{pano2020complete} researched if outbreaks of COVID-19 can influence Bitcoin prices. They performed 13 different strategies for BTC tweets. VADER scoring systems are regarded as the optimum processing approach. Mohapatra et al.\cite{mohapatra2019kryptooracle} had attempted to assign each tweet a compound sentiment score based on the VADER sentiment analysis algorithm. The number of Twitter followers, number of likes, and number of retweets associated with each tweet is used for the final sentiment score. 


Three-level hierarchical Bayesian model, \emph{Latent Dirichlet Allocation}(LDA), is generative probabilistic model for finding patterns of words in text corpus\cite{blei2003latent}. LDA is demonstrated that it outperforms batch variational bayes (VB) and also need less running time\cite{hoffman2010online}. The performance of classical state space models and specify a statistical model of topic evolution has been enhanced by David et al.\cite{blei2006dynamic}. Based on probabilistic time series, this dynamic model can capture the evolution of topics in a corpus. One of LDA limitations is the incompetence to model topic correlation. \cite{blei2006correlated} has presented the \emph{correlated topic model}(CTM) with respect to this limitation. The CTM directly models correlation between topics by using co-variance structure among the components. This proved correlation play an important role in topic modelling. In\cite{newman2010automatic}, key words are used to represent topics. Automatic coherence evaluation was proposed to rate coherence or interpretability. Michael et al.\cite{roder2015exploring} proposed a framework that combines existing word-based coherence measures and the combinations of basic components. This configuration space has explored the best overall correlation for the coherence definition with respect to all available human topic ranking data.

%
\section{Methodology}

\subsection{Overview}
Firstly, we collect the tweets data from a public Twitter dataset. It contains millions of tweets data related to COVID-19. For this dataset, we only retain the data from the UK or US and related to the vaccine. After obtaining the data, it is necessary to execute data processing to remove redundant and invalid content appearing in the tweet text. In the analyzing steps, the whole dataset is split differently depending on the requirements of three different questions. During the experiments, we mainly apply VADER to implement the sentiment analysis. Besides, high-frequency words are collected and displayed by word cloud. And the LDA model is applied to three aspects, countries, emotion, and time series. Subsequently, we model and analyze the data from different countries, sentiments and periods to dig the potential attitude towards the vaccines and coronavirus on Twitter.

%

\subsection{Data collection}
Banda et al.\cite{banda2020large} provides a dataset of twitter posts in four languages about the COVID-19 vaccine. This dataset is updated at least once every fortnight. Therefore, based on this public dataset, we further collected vaccine-related tweets posted by residents of the UK and US during the outbreak.

\begin{figure*}[htbp]
    \caption{The steps of dataset-collecting. According to the public dataset and first multiprocessing, the residents' tweets in the GB and UA are maintained. Subsequently, the fields of specific tweets are collected by using Twitter APIs and the second multiprocessing. Finally, we get final dataset for further research. }
    \centering
    \includegraphics[width=\linewidth]{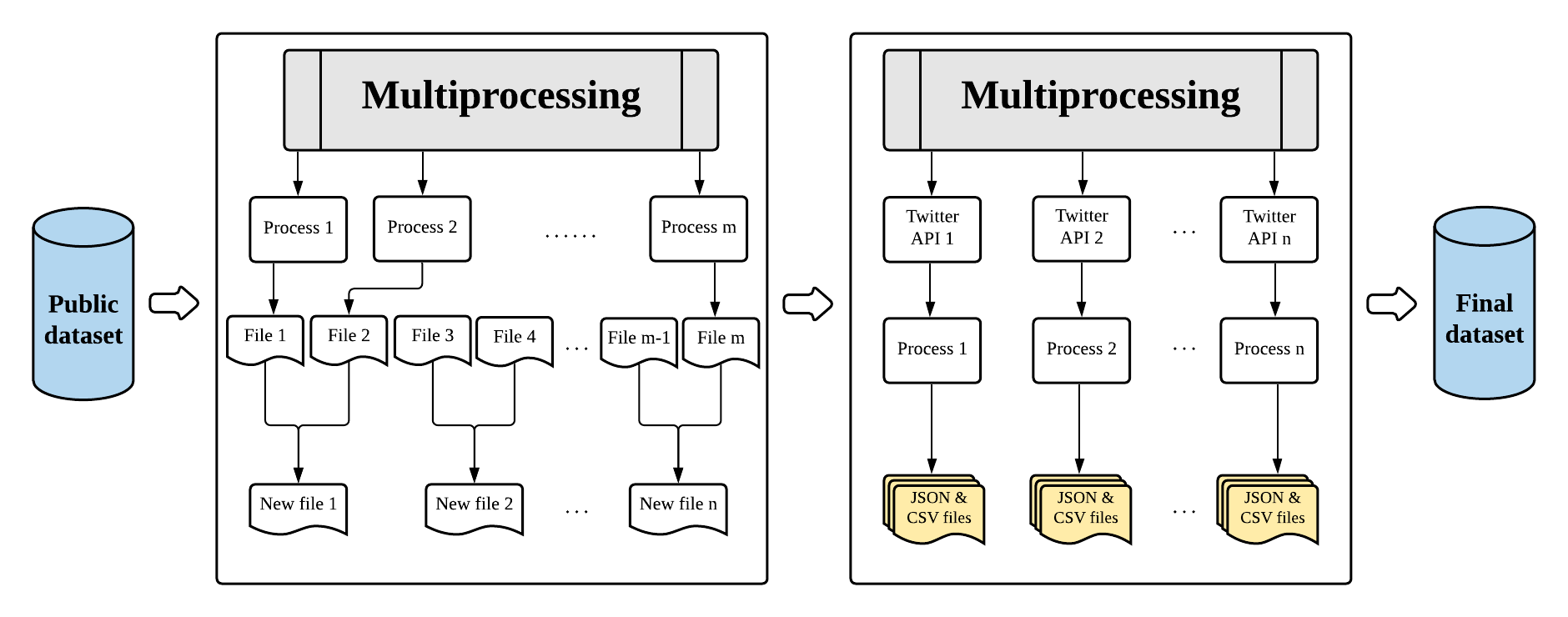}
    \label{fig:data_collection}
\end{figure*}

The specific dataset acquisition process is defined as following (Fig.\ref{fig:data_collection}):
\begin{itemize}
  \item The public dataset file\cite{banda2020large} is downloaded.
  \item The path list for 432 days' dataset is created for later use.
  \item According to the field, country\_place in the public dataset, the tweets were posted by residents in the GB and US are retained.
  \item Multiprocessing features are used to speed up the data processing. Specifically, 8 processes are used to do that.
  \item We combine the 8 data results obtained from the last step to generate 4 data sets for later use.
  \item We use the multiprocessing technique to set 4 processes for collecting twitter posts or tweets. Based on the field of tweet\_id in the public dataset obtained from the last step, we acquire vaccine-related tweet posts via four Twitter APIs in parallel.
  \item Only the 15 key fields are kept in the 4 separate files for each tweet post. Finally, those 4 files are converted into 4 CSV files, and then, they are merged for subsequent research.
\end{itemize}

Our dataset retains the "retweet\_count" field. This field can help us track and research important tweets. In addition, every tweet we collected is fully completed(including the memes in the tweets). Through the above method, a dataset of about 110,000 British and American tweets from 25 January 2020 to 14 March 2021 is collected. For this dataset, the next stage of data pre-processing is carried out to ensure the accuracy of the dataset during the research. The relevant changes of this dataset are continuously monitored by our team members during the study.

\subsection{Data pre-processing}
Since the Twitter text data has some typical but non-semantic features, we processed these features in the data pre-processing steps.

First, almost every tweet contains a short link, it can be "https://t.co/o7amgl8ybl", this kind of link do not have the actual semantics and can cause ambiguity after tokenization, such as "o7amg" is divided into "o", "7" and "amg", and "amg" could be taken as a car brand. This causes unnecessary trouble to later analysis, so remove of this link is needed.
Second, although VADER can score punctuation, it is only used by an exclamation point (!), so other punctuation is deleted in this step.
Third, we remove unnecessary line breaks because this is the most common meaningless identifier.
Fourth, conversion of all the text content to lowercase is conducted to ensure every word appears in a consistent format.

\subsection{Sentiment analysis}
For the first and second questions we raised, we try to use sentiment analysis to understand the attitudes of American and British citizens towards COVID-19 vaccine from the perspective of emotional expression. At the same time, we used multiple dimensions to try to explain people's acceptance of different vaccine brands, and how attitudes towards vaccines have changed over time.

In our study, sentiment analysis is mainly realized through VADER. This tool's sentiment lexicon is sensitive towards the polarity and intensity of subjectivity expressed in social media contexts and is also widely applicable to other domains.

Generally speaking, sentiment analysis mainly determines the proportion of positive, negative and neutral texts (the proportion of polarity) and the intensity of their emotional expression in a given text through various methods. Finally, a predefined rule is used to make a category judgment or a comprehensive score for the text.

VADER is a gold-standard sentiment lexicon obtained through the Wisdom of the Crowd (WotC) approach. Through extensive human work, the emergence of this dictionary enables the emotional analysis of social networks to be completed quickly and has a very high accuracy comparable to that of human beings\cite{hutto2014vader}.
As a complete tool, we can use VADER to obtain sentiment scores at the document level, sentence level, and phrase level, depending on the granularity of the analysis for the actual application scenario.

In the scenario we studied, the minimum unit of analysis was set to the preprocessed text data of a tweet. At the same time, according to the classification method recommended by Hutto et al., we mapped the emotional score into three categories: positive, negative and neutral through the same parameter setting.

Interestingly, based on the rich data returned from the Twitter API, we were able to capture a lot of non-text data in this study, such as a user's tweeting location, country code, time, follower number, likes number, retweet number, listed number(how many lists the account is contained in). Based on these data, our analysis can be carried out in terms of geographical region and time.

The study was conducted on all tweets in English, which would include tweets from many English-speaking countries. Obviously, if we do not make a distinction between the regions where people tweet, it is easy to get insignificant results. Because the epidemic situation in different countries is affected by the national governance ability, economic level, scientific and technological level, international political status and many other aspects, the vaccination opportunities of people in different countries are greatly different. Meanwhile, due to international politics, different countries promote different vaccine brands (Chinese vaccine, American vaccine, British vaccine, etc.). The effectiveness and safety of different vaccines also vary. In other words, common sense suggests that the severity of the epidemic in a country and the speed of vaccination, the brand of the vaccine, have a big impact on the acceptance of the vaccine.

Therefore, we selected the two most representative English-using countries, the United Kingdom and the United States, and analyzed them respectively.

In terms of time, considering the imbalance of data volume and data distribution, we divide the data according to month as the smallest unit. The number and proportion of tweets in different categories, as well as the average emotional score of all tweets in each month, were calculated.

As far as we know, existing studies on sentiment analysis of COVID-19 have ignored the dynamics of opinion transmission in social networks \cite{kempe2003maximizing}. In other words, all users are regarded as individuals with the same influence in analysis so as to reach the final conclusion. Such a method deviates from the actual sentiment.

After considering different diffusion models\cite{kempe2003maximizing}, we believe that follower number and listed number have the same nature of degree in social network, and retweet number and likes number reflect the potential influence of a single user. That is, the degree is directly proportional to the follower number and listed number, which means that the user is in contact with more users and can impose influence on them. Retweet number and likes number are directly proportional to the potential impact, and these two metrics can be interpreted as content quality metrics. Therefore, we propose the following improvements to the compound score:

\begin{center}
$Final compound= compound$
$*(retweet number + likes number)$
$*(follower number + listed number)$
\end{center}

The Final compound above is used for the study of question one, and we name the method as "Weighted-VADER".

\begin{figure}[htbp]
    \centering
    \caption{Dynamic topic modelling. A set of topics in the dataset are evolved from the set of the previous slice over time series.}
    \includegraphics[width=\linewidth]{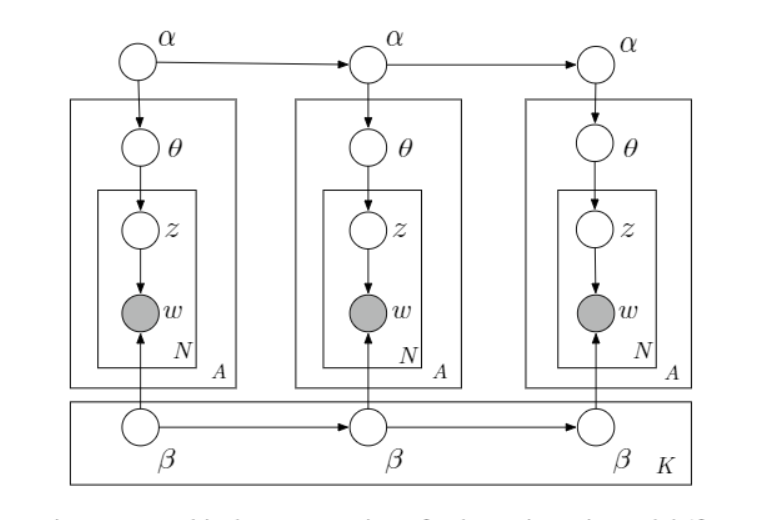}
    \label{fig:dtm}
\end{figure}

\subsection{LDA}
As for the third question, we choose the LDA model to build topics. The LAD model can analyze representative and valuable objects from tweets. An effective LDA model can generate a predefined probabilistic procedure\cite{hoffman2010online}. There are the basic processing steps. The first step is to choose a distribution over a mixture of K topics. Next, it selects a topic and draws a word from that topic by the topic’s word probability distribution. Finally, it produces a K-topic list ranked by the percentage of the total number of words related to this topic. Each topic shows the most relevant words.

Meanwhile, we find the content on Tweet is changeable, and the topic content is not a static corpus, which spans over a couple of months. The dynamic topic model is a generative model for analyzing topic evolutions in a gigantic corpus\cite{blei2006dynamic}. As a part of the probabilistic topic models class, the dynamic one can catch how various themes on tweets evolved. The whole period is split into several time-slices. These time-slices are put into the model provided by "gensims"\cite{rehurek_lrec}. The details about the DTM are illustrated in Fig.\ref{fig:dtm}.

An essential challenge is to determine an appropriate topic number for the topic model. Michael et al.\cite{roder2015exploring} proposed coherence score to evaluate the quality of each topic model. Coherence scores measure the consistency of the words that compose a topic. Besides the distribution on the primer component analysis (PCA) is considered, which can visualize the topic models in a word spatial with two dimensions. A uniform distribution is preferred, which is considered as a high degree of independence for each topic. The judgement for a good model is a higher coherence and an average distribution on the primer analysis displayed by the pyLDAvis\cite{sievert2014ldavis}. The first point indicates that the content within each topic is highly harmonized. The second point means the model has fewer intersections among topics, which summarizes the whole word space well and remain relatively independent.

%
\begin{figure*}[htbp]
    \centering
    \caption{The total text word frequency. After removing irrelevant word, word frequency of main opinion is counted and visualized.}
    \includegraphics[width=\linewidth]{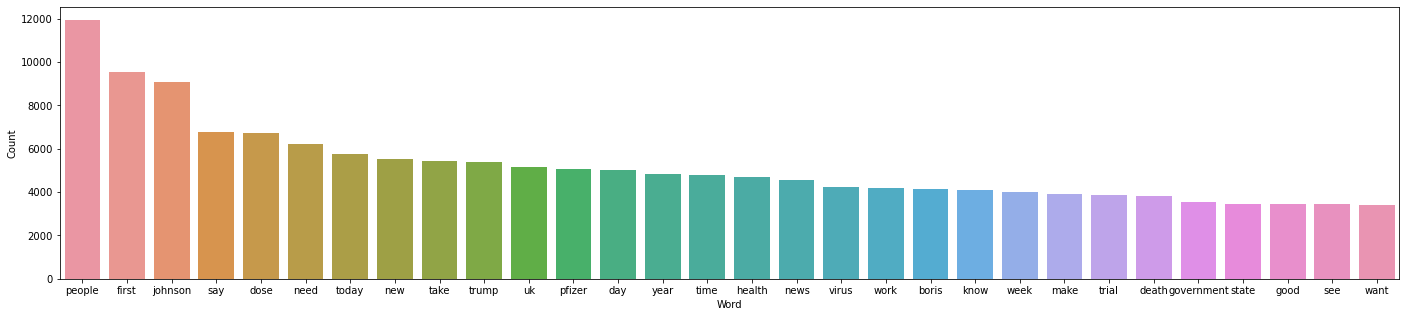}
    \label{fig:total_text_word_frequency}
\end{figure*}

\section{Experiments}

\subsection{Dominant sentiment towards COVID-19 vaccines} \label{Q1}

The sentiment analysis is applied to section \ref{Q1} and section \ref{Q2}.  

We first filtered the data by location to get all the tweets with "country code" fields "USA", "GB" and then divided it into two subsets, each of which was further divided by month.

To classify the sentiment of a tweet as positive, negative or neutral one, we follow the classification method proposed in \cite{hutto2014vader}. After executing the VADER tool on a tweet, it generates four scores including "pos", "neg", "neu" and "compound". The "compound" score is calculated based on specific rules and the valence score of each word in this tweet. It is also normalized to ensure the ranges from -1 to 1. Therefore, the "compound" score is considered to represent the sentiment of this tweet. By setting the classification threshold for the "compound" score, the sentiment of a tweet can be determined. Depending on the rule from \cite{hutto2014vader}, the threshold values are -1, -0.05, 0.05 and 1. When the score is greater than 1 and less than or equal to -0.05, this tweet is considered a negative tweet. When the score is greater than or equal to 0.05 and less than 1, the sentiment of this tweet is determined as positive. For the tweet with a score in the range from -0.05 to 0.05, it is considered a neutral tweet.

Weighted-VADER and VADER methods are used to conduct a control experiment on the processed dataset.

\subsection{Sentiment analysis for different vaccine brands} \label{Q2}

By executing the data cleaning process, the tweets text data with less confusion can be obtained. With respect to this particular experiment, the processed dataset is separated into two based on the locale (the UK or US) of tweets. The duplicate tweets are removed and only the unique tweets remain. For tweets data in the UK and US respectively, we analyze which vaccine brands are talked about more frequently and how the opinions of people toward them.

To determine the vaccine brands we would analyze, we browsed many news and resources on social media platform related to vaccine brands. We found that there are 12 large enterprises that currently make great progress in researching and manufacturing the COVID-19 vaccine. They are Sinopharm, Sinovac, Cansinobio, Novavax, AstraZeneca, Johnson \& Johnson (Janssen), Sanofi \& GlaxoSmithKline, Moderna, Pfizer, Sputnik-V, Valneva and CureVac. The vaccine from these brands is already widely in use or ready to be used soon. We investigated what corresponding keywords are used most frequently on Twitter for each vaccine brand. The corresponding keywords of a certain vaccine brand would be highly possible to be included in tweets when a tweet talks about this brand. For instance, when users talk about Johnson \& Johnson on Twitter, it is most likely keywords such as "johnson \& johnson", "johnsonjohnson", "jnjnews", "janssen" and "johnsonandjohnson" occur in the tweet text.

For different vaccine brands, we counted the number of tweets that talk about them respectively. It aims to analyze which brand is the most popular and widely known in the UK or US. Furthermore, we applied VADER to implement the sentiment analysis for tweets discussing different brands. Afterwards, the tweets' number of each sentiment class (positive, negative, neutral) would be recorded. The purpose is to research people's opinions regarding each vaccine brand. It can also help to realize that if people in the UK or US prefer to or dislike a certain brand.

\subsection{Extraction of the mainstream words and opinion } \label{Q3}
In this part, a higher quality dataset is required for topic model. Apart from the general data cleaning methods, lemmatization could enable the model to achieve better performance. The different forms of a word cause the misclassification for models. Consequently, NLTK\cite{bird2009natural} is used to accomplish lemmatization. Finally, We pruned the vocabulary by stemming each term to its root, removing stop words, and removing terms that unrelated to the topic like "wa", "ha" and "would". The total size of vocabulary is 41645 in this part. After the data processing, we counted word frequency and displayed the top 30 words in Fig.\ref{fig:total_text_word_frequency}. And the word cloud was generated to analyze the main opinion.

\begin{figure*}
    \centering
    \caption{Coherence values. The topic coherence is the measure to evaluate the coherence between topics inferred by a model. Left: based on a sliding window, the cv measure uses normalized point-wise mutual information (NPMI) and the cosine similarity. Right: according to document co-occurrence counts, the umass evaluation is confirmed by segmentation and a logarithmic conditional probability.}
    \includegraphics[width=\linewidth]{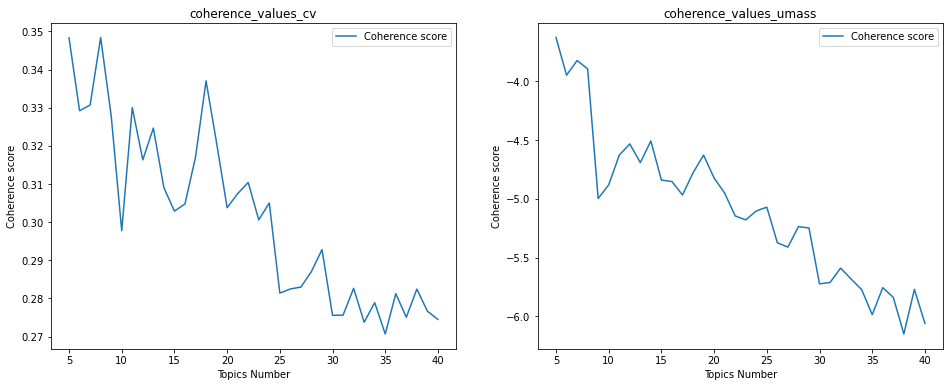}
    \label{fig:coherence_values}
\end{figure*}

\begin{figure*}[htbp]
    \centering
    \caption{Primer component analysis. The average distribution is used for primer component analysis. Left: the coherence score reached at the highest point at eight topic numbers. But the coherence scores fluctuate sharply. Right: 18 is the final choice for our topic number, which is a local maximum of 8-40.}
    \begin{subfigure}[t]{0.45\linewidth}
        \centering
        \includegraphics[width=\linewidth]{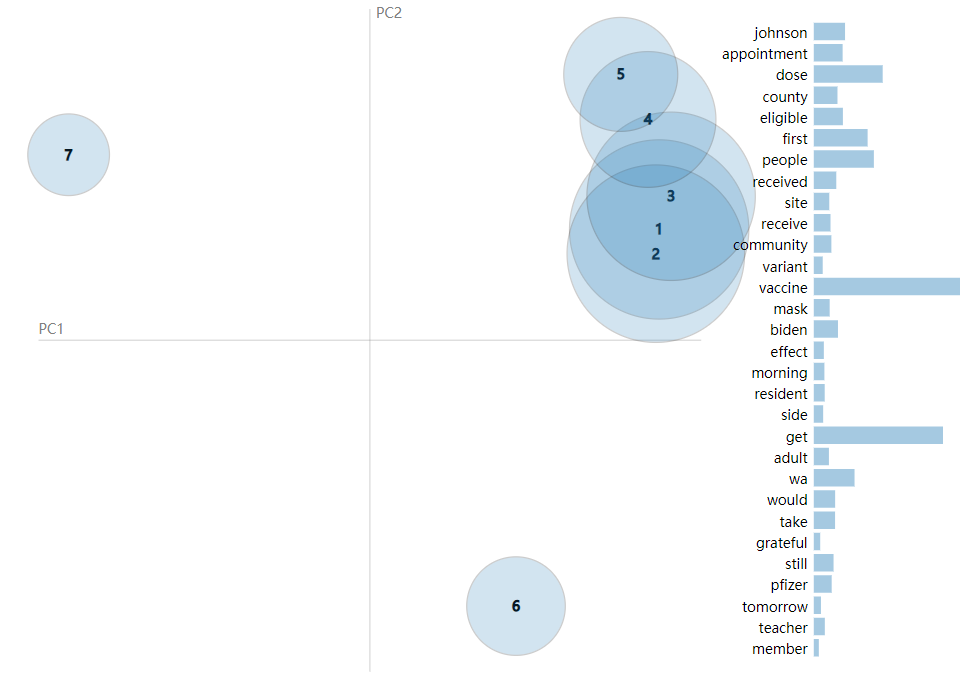}
        \caption{Topic numbers= 8}
        \label{fig:pca1}
    \end{subfigure}
    \begin{subfigure}[t]{0.05\linewidth}
        \centering
        \includegraphics[width=\linewidth]{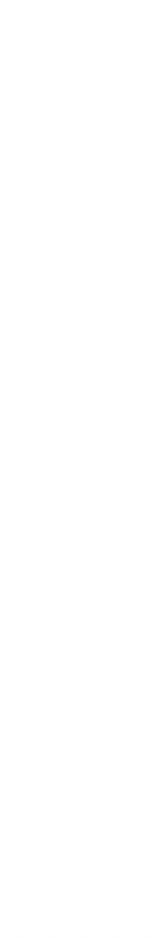}
    \end{subfigure}
    \begin{subfigure}[t]{0.45\linewidth}
        \centering
        \includegraphics[width=\linewidth]{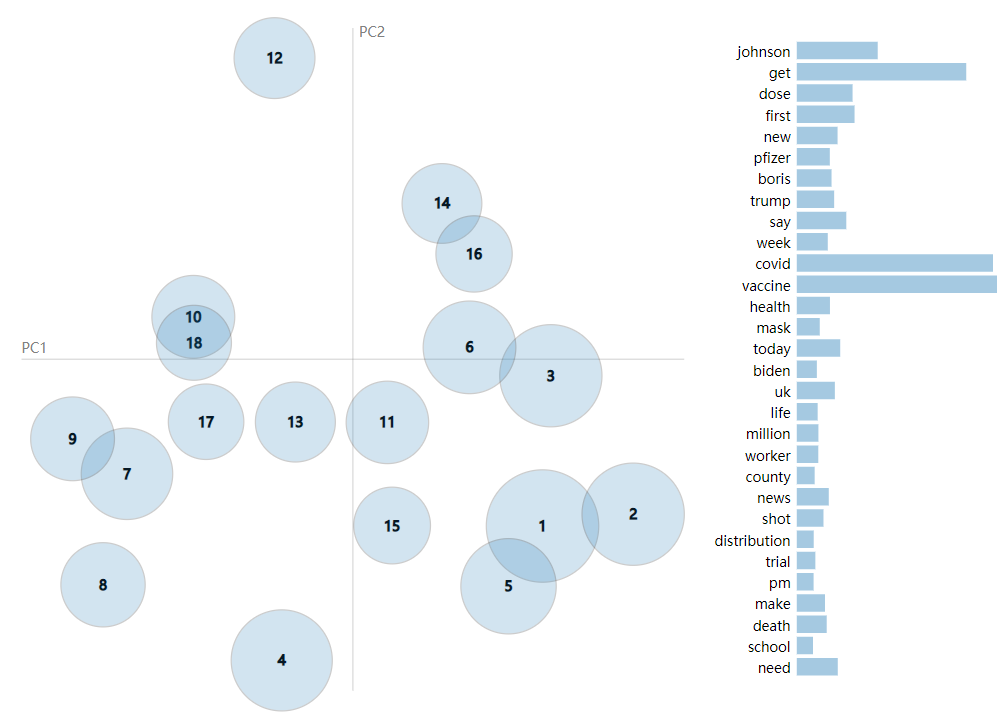}
        \caption{Topic numbers= 18}
        \label{fig:pca2}
    \end{subfigure}
    
    \label{fig:pca}
\end{figure*}

To explore what the user concerns about on Twitter, we applied the LDA to our clean corpus. And for a better representation of the whole content, it is necessary to find an appropriate topic number. By using the topic number ranging from 5-40, we initiated the LDA models and calculated the model's coherence. According to Fig.\ref{fig:coherence_values}, the coherence score peaked at eight topic numbers. However, an unexpected aggregation of topics appears in Fig.\ref{fig:pca1}. We mainly used "cv" coherence as an indicator and "umass" coherence as a secondary reference. Coherence scores showed an overall downward trend but still fluctuated considerably. Consequently, an accepted method is to choose the local maximum with the average distribution. Finally, we chose 18 as the topic number (Fig.\ref{fig:pca2}), which reaches the highest point from 8-40.

Besides, due to the complexity of the entire tweet, the information obtained is very general. To analyze more precisely the attitudes towards vaccines in different periods, countries, and emotions, we divided the dataset and modeled them from the following aspects:
\begin{itemize}
  \item Time Series analysis: We assumed the topics change slightly over time, which is suitable for dynamic topic modelling. Therefore, the whole dataset was split into three time-slices to feed the dynamic topic model. And we explored the word changes in each topic.
  \item Country analysis: To investigate the different concerns between the UK and US, we divided tweets into two parts according to their countries and analyzed them separately using LDA model.
  \item Sentiment analysis: Because different emotions express different themes, we divided the dataset into positive and negative subsets and performed LDA analysis on each of them to infer the effect of the vaccine.
\end{itemize}


\subsection{Experiment results}

\begin{figure*}[htbp]
    \centering
    \caption{Sentiment analysis by two models on the USA and GB Twitter text data. The four result graphs were all composed of three sub-graphs: Tweets Number by label, Tweets ratio by label and Average compound score. The left column shows results performed by VADER, right column displays outcomes executed by Weighted-VADER, where the Number of Tweets is the Number after weight adjustment, so its value is much larger than the normal.}
    \begin{subfigure}[t]{0.48\linewidth}
        \centering
        \includegraphics[width=\linewidth]{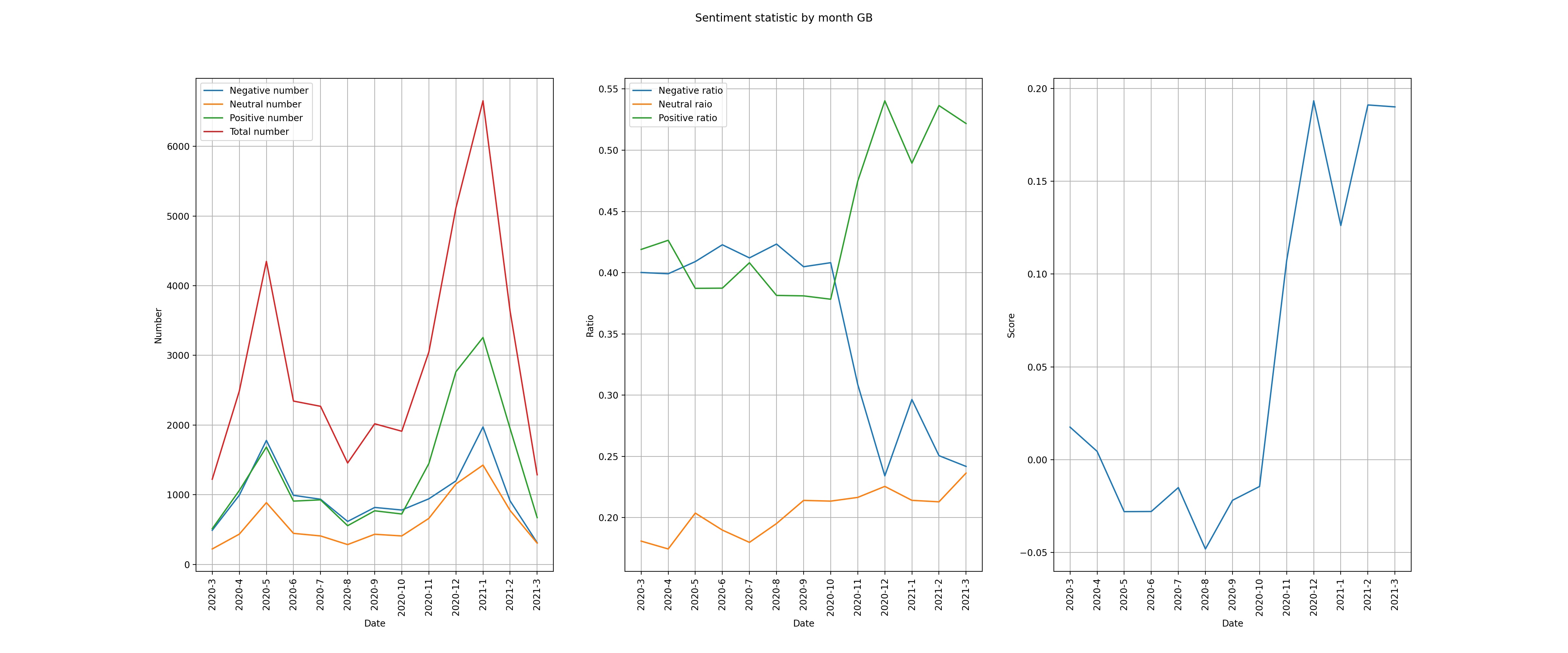}
        \caption{SMA on GB tweets by month}
        \label{fig:SMA_GB}
    \end{subfigure}
    \begin{subfigure}[t]{0.48\linewidth}
        \centering
        \includegraphics[width=\linewidth]{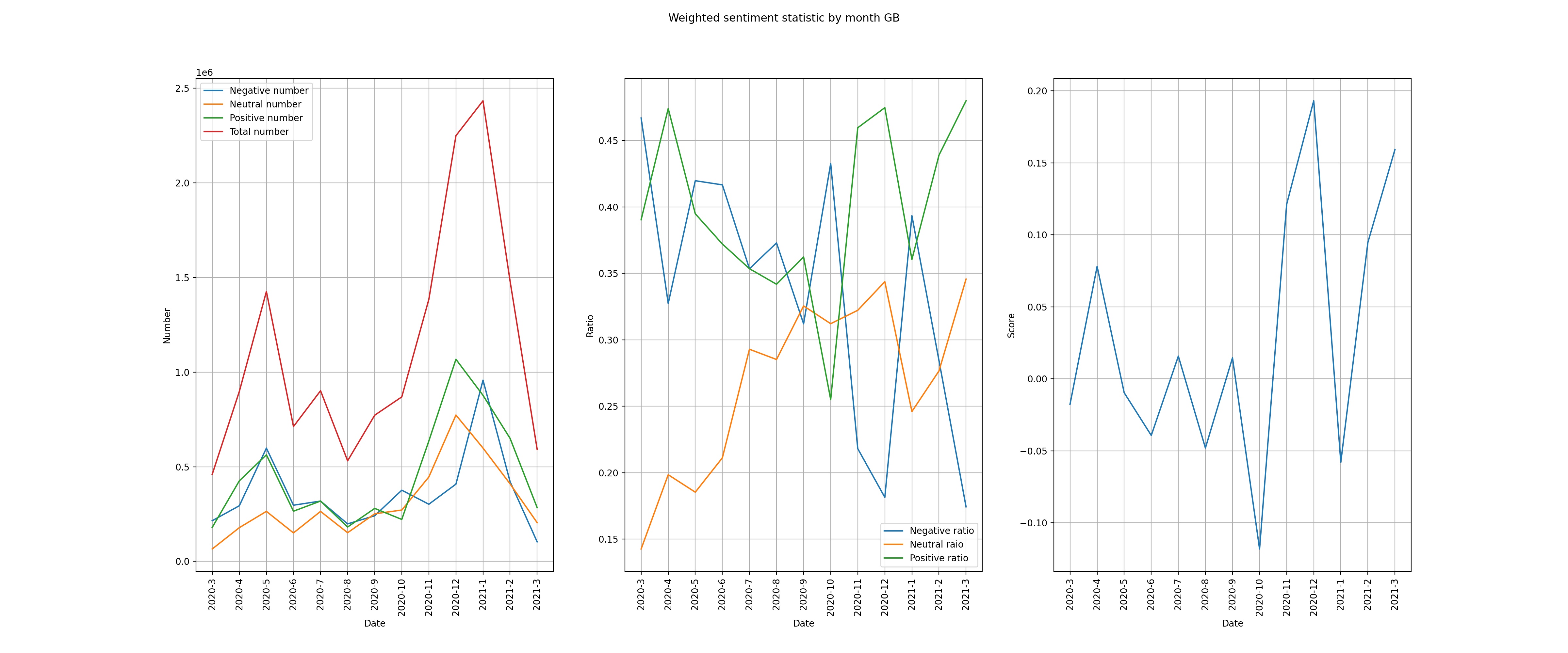}
        \caption{Weighted SMA on GB tweets by month}
        \label{fig:Weighted_SMA_GB}
    \end{subfigure}
    \begin{subfigure}[t]{0.48\linewidth}
        \centering
        \includegraphics[width=\linewidth]{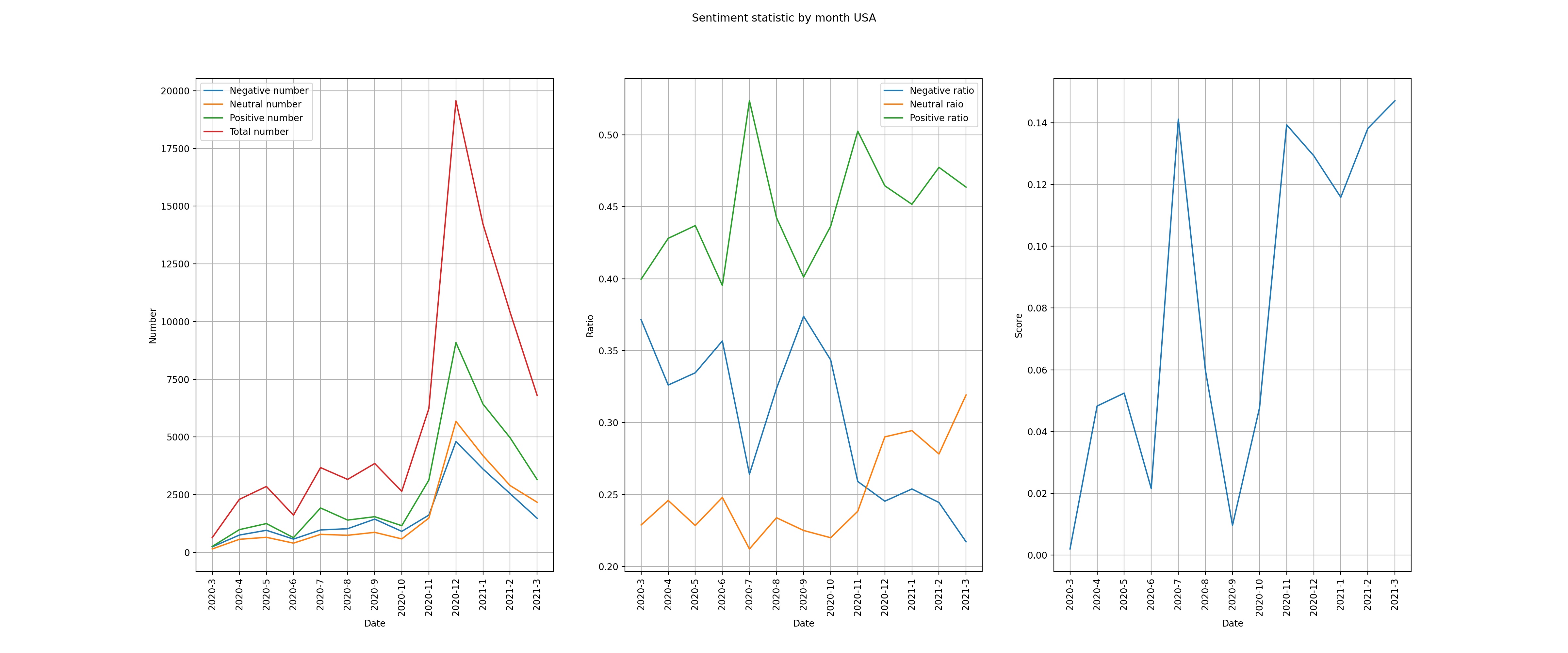}
        \caption{SMA on the USA tweets by month}
        \label{fig:SMA_USA}
    \end{subfigure}
    \begin{subfigure}[t]{0.48\linewidth}
        \centering
        \includegraphics[width=\linewidth]{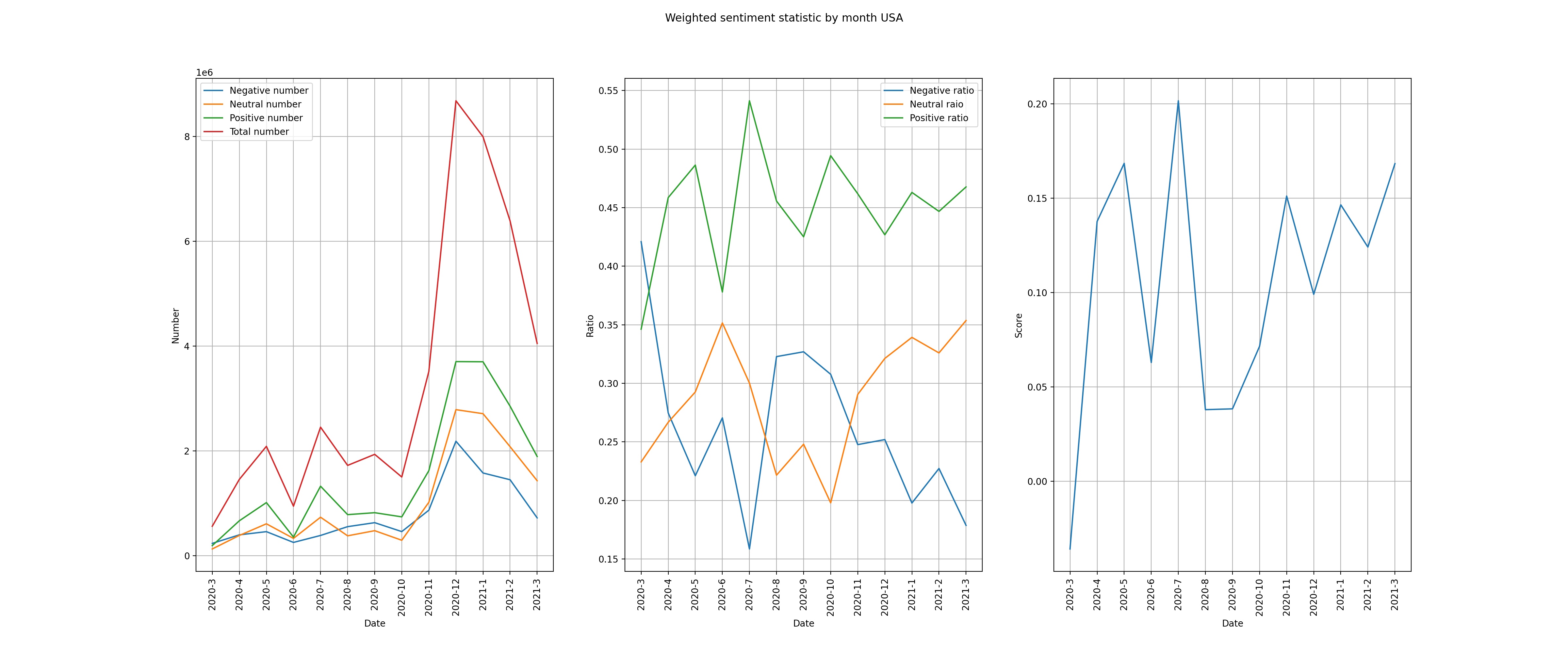}
        \caption{Weighted SMA on the USA tweets by month}
        \label{fig:Weighted_SMA_USA}
    \end{subfigure}
    \label{fig:SMA}
\end{figure*}

Although the data is not evenly distributed in each month, we believe it is enough to produce effective analysis results.
This is because the current data is randomly sampled from Twitter, so the distribution of the dataset at different times can be considered to be the same as the actual distribution. According to the statistics of Fig.\ref{fig:SMA_GB} and \ref{fig:SMA_USA}, it shows that the cumulative number of tweets has increased since the beginning of the pandemic. In the plot, the US has a significant peak in December 2020, while the UK has a peak in May 2020 and January 2021. However, the US did not have this peak at the beginning of the epidemic, which may be due to the government's cover-up and neglect of the COVID-19 epidemic, as well as the influence of the anti-vaccination movement in the US \cite{larson2020blocking}.

In Fig.\ref{fig:SMA_GB} and \ref{fig:SMA_USA}, we can also observe that only the United States consistently has a more significant percentage of positive tweets than tweets with other attitudes in all months.
Notably, approximately 40\% of the population in both countries hold a negative attitude towards COVID-19 vaccines by October 2020.
That is not a good sign of the acceptance of vaccines.
As the vaccination in December 2020 started \footnote{https://ourworldindata.org/covid-vaccinations}, 20\% of the decrease appeared in the UK, but only 10\% of the decline happened in the US. The government should focus on the cause of the negative attitude towards the vaccine, and formulate the corresponding propaganda and education policies, make the mass vaccination finish faster.
As for the results presented by the Weight-VADER model in Fig.\ref{fig:Weighted_SMA_GB} and \ref{fig:Weighted_SMA_USA}, it can be seen from the comparison with Fig.\ref{fig:SMA_GB} and \ref{fig:SMA_USA} that under the influence of these more influential users (the peak value in the Ratio-Date graph), all the turning points of the proportion of users' different opinions in unweighted sentiment analysis can be well explained.
In other words, in the context of vaccine promotion, people with public influence should play a positive role in vaccination promotion.
In Fig.\ref{fig:Weighted_SMA_GB} (Number-date), we can see a substantial positive peak in December 2020, which we believe is due to some influential official accounts' publicity, resulting in a sharp rise in the overall weighted number.
In the Fig.\ref{fig:SMA_GB} and \ref{fig:SMA_USA} (Score-Date), the overall sentiment is not apparent, but in the Fig.\ref{fig:Weighted_SMA_GB} and \ref{fig:Weighted_SMA_USA}, different trends can be observed. The British people's attitude towards vaccines changed in a more dramatically way(-0.55 to 0.75), compared with the American people's (-0.3 to 0.55).


Regarding to the analysis about vaccine brands, the experiment result can be observed in Table \ref{tab:Q2 result}. From Table \ref{tab:Q2 UK result}, it can be seen that for the tweet data in this particular dataset, nearly 55\% and 30\% of tweets related to COVID-19 vaccine talk about Pfizer and AstraZeneca respectively. It demonstrates that these two brands are the most popular in the UK. From the result in Table \ref{tab:Q2 US result}, we can see Pfizer and Moderna are the most popular vaccine brands in the US, there are approximately 57\% and 31\% of tweets talking about them respectively. Whatever in the UK or US, Pfizer is the most popular vaccine brand.

In terms of the sentiment analysis, we can see a majority of tweets express positive and neutral opinions toward vaccine brands. Furthermore, the proportion of positive tweets is greater than the neutral tweets for most brands. It shows that in the UK and US, the attitude of most people toward different brands of COVID-19 vaccine is positive.

\begin{table*}
\centering
\caption{Sentiment analysis result of tweets toward vaccine brands. Columns from left to right refer to name of vaccine brands, the number of tweets talking about the corresponding brand, the number of positive, neutral, negative tweets among all tweets of this brand, and the percentage of the number of tweets for this brand in the total.} 

\begin{subtable}{1\textwidth}
\centering
\begin{tabular}{cccccc} \hline
Vaccine Brand & Number of Tweets & Positive Tweets & Neutral Tweets & Negative Tweets & Percentage of total tweets \\ \hline
Sinopharm & 9 & 5 & 0 & 4 & 0.00416 \\
Sinovac & 12 & 8 & 2 & 2 & 0.00555 \\
Novavax & 96 & 74 & 10 & 12 & 0.04438 \\
AstraZeneca & 666 & 421 & 143 & 102 & 0.30791 \\
Johnson \& Johnson & 15 & 10 & 3 & 2 & 0.00693 \\
Sanofi & 30 & 19 & 6 & 5 & 0.01387 \\
Moderna & 128 & 81 & 35 & 12 & 0.05918 \\
Pfizer & 1182 & 701 & 258 & 223 & 0.54646 \\
SputnikV & 9 & 4 & 0 & 5 & 0.00416 \\
Valneva & 9 & 5 & 4 & 0 & 0.00416 \\
CureVac & 7 & 5 & 1 & 1 & 0.00324 \\ \hline
\end{tabular}
\caption{UK result}
		\label{tab:Q2 UK result}
\end{subtable}%

\begin{subtable}{1\textwidth}
\centering
\begin{tabular}{cccccc} \hline
Vaccine Brand & Number of Tweets & Positive Tweets & Neutral Tweets & Negative Tweets & Percentage of total tweets \\ \hline
Sinopharm & 19 & 12 & 4 & 3 & 0.00242 \\
Sinovac & 27 & 13 & 12 & 2 & 0.00343 \\
Novavax & 148 & 86 & 45 & 17 & 0.01882 \\
AstraZeneca & 497 & 219 & 137 & 141 & 0.06319 \\
Johnson \& Johnson & 177 & 85 & 57 & 35 & 0.0225 \\
Sanofi & 50 & 32 & 9 & 9 & 0.00636 \\
Moderna & 2386 & 1384 & 642 & 360 & 0.30337 \\
Pfizer & 4535 & 2398 & 1214 & 923 & 0.57661 \\
SputnikV & 17 & 6 & 4 & 7 & 0.00216 \\
CureVac & 9 & 7 & 1 & 1 & 0.00114 \\ \hline
\end{tabular} 
\caption{US result}
		\label{tab:Q2 US result}
\end{subtable}
\label{tab:Q2 result}
\end{table*}

\begin{table*}[]
\centering
\caption{The LDA investigation for the whole text. In terms of percentage size, the vaccine-relative topics are ranked and listed. Also, the first eight words of each topic are recorded in the table.}
\begin{tabular}{ccccccccc}
\hline
Topic number     & Word 1  & Word 2  & Word 3       & Word 4 & Word 5   & Word 6 & Word 7  & Word 8     \\ \hline
Topic 2 (7.5\%)  & vaccine & get     & need         & need   & make     & know   & safe    & want       \\
Topic 3 (7.5\%)  & vaccine & covid   & help         & people & immunity & time   & virus   & well       \\
Topic 5 (6.5\%)  & get     & vaccine & week         & worker & covid    & care   & people  & healthcare \\
Topic 8 (5.1\%)  & vaccine & covid   & pfizer       & does   & million  & news   & moderna & company    \\
Topic 16 (4.2\%) & life    & vaccine & distribution & world  & coming   & covid  & save    & leader    \\ \hline
\end{tabular}
\label{tab:Q3 result}
\end{table*}

The word frequency and word cloud statistics for the entire dataset are shown in Fig.\ref{fig:total_text_word_frequency} and Fig.\ref{fig:pca} respectively. We noticed words frequency, like "people", "get", "first" and "dose", are at the top of the list. This means the majority of residents are willing to share their vaccination status on social media platforms. Moreover, citizens are worried about governments' policies concerning COVID-19. The heads of state in both countries are mentioned frequently on vaccine topics.

The LDA analysis for the whole text is shown in Table \ref{tab:Q3 result}. The table shows that the second and third themes take up 15\% of the total tokens, including the words "token", "people", "vaccine", "immunity" and "safety". Based on this, we inferred that most people accept vaccination because they believe it has positive effects. Besides, several words, like "medical", "worker", "get" and "vaccine", are mentioned in the fifth topic. This indicates that healthcare workers should be the first group to be vaccinated. As for the eighth topic, Pfizer, Modena and AstraZeneca are frequently reported in the UK news. So, our analysis of the experimental results is accurate.

The dynamic topic model has shown that the people's concern remain stable over the period. Words such as "get", "worker" and "health" gained more importance in the later stages of the pandemic. While the word "distribution" ranked low at first but started climbing in the middle of the period. Those results indicated that there is growing concern about the distribution of vaccines.

Analyzing from the country perspective, we found that British people worried about the vaccine shortage. Nevertheless, American are keen to talk about vaccination. Most US residents have no adverse reactions after vaccinating, and they have faith in vaccination. Similarly, both countries' citizens are greatly concerned about governments' vaccine allocation.

By analyzing the positive sentiment, nearly 9\% of tokens consider the vaccine to be effective, ranking first in this model. Furthermore, 8.5\% of tokens mention “free” and “safe” after the vaccination. By analyzing the negative topic model, words like "death", "case" and "news" appear in the list of topics 2, which could be interpreted as the news of vaccination deaths has caused people's anxiety towards COVID-19 vaccine. The third topic reflects that the lack of vaccine is still a severe problem. The rest of the negative thoughts mainly come from irrelevant COVID-19 vaccine events such as lockdown, mask-wearing and virus-damaging.

\section{Limitation}
The research dataset in our experiment is completely insufficient. Based on the UK and US residents' tweet posts obtained via the tweet API, we only kept tweets including the keyword "vaccine". If more research time were available, we would like to implement cross-platform data research, such as combining Instagram, YouTube and Twitter. We will use multi-modal sentiment analysis methodology to implement sentiment analysis, including the analysis of text, audio and video simultaneously. If we use the multi-modal method, we can get more comprehensive and accurate experimental results of people's attitudes towards vaccines.

Social media have the characteristic of high interaction, rapid spread and immediate change. Moreover, Twitter content is indeterminate. The hashtag and keyword of COVID-19 vaccine can change at any time depending on the epidemic and the vaccine development. Therefore, our research results about this topic are time-sensitive. It is only relevant for the tweet posts during our research period.

We have not further improved the sentiment analysis and LDA model. The experimental model is based on existing libraries. If we could have a chance, we would try to compare the accuracy of different models. We will attempt to enhance the accuracy of the sentiment analysis and the topic modelling.

\section{Conclusion}\label{AA}
In conclusion, it is important to analyze individuals’ sentiments towards COVID-19 vaccine by their social media comments during the dramatic outbreak of the coronavirus disease. To analyze dwellers attitude towards COVID-19 vaccine in the UK and US, we designed and executed a series of experiments on the dataset collected from Twitter. 
We found that through the VADER and Weighted-VADER methods, a quantitative and qualitative research can be conducted. Public attitudes towards vaccines have improved sharply following the rapid progress of vaccination in the UK and the US, but there is still a significant proportion of people with negative attitudes.
Besides, among different brands of COVID-19 vaccines, Pfizer is the most talked about one in the UK and US. Most citizens hold a positive view of the COVID-19 vaccine.
Through the LDA analysis, we found that most people hope to be vaccinated and feel great after receiving the vaccine. Public concerns about vaccines stem mainly from death cases of vaccinated people. For the rest with negative attitudes, few people criticize the vaccine directly. They merely mention the vaccine when complaining about the damage caused by the virus. Other concern topics are vaccines distribution, the relationship between schools and vaccines, and vaccines appreciation.

In future work, we will redesign the functions within the VADER tool for sentiment analysis. With new functions, we will further improve the accuracy of sentiment recognition. Meantime, our team will collect more tweet posts for model training. We will also do further research on the LDA model.

\bibliographystyle{IEEEtran}
\nocite{*} 

%
\bibliography{references}

\end{document}